\documentclass[letter,10pt,conference]{style/ieeeconf}
\IEEEoverridecommandlockouts
\usepackage{amsmath,amsfonts,amssymb}
\usepackage{graphicx}
\usepackage{gensymb}
\usepackage[ruled]{algorithm}
\usepackage{algpseudocode,caption}

\begin{document}

\title{\LARGE \bf
\vspace*{-0.25em}
Path-aware optimistic optimization for a mobile robot
\vspace*{-0.5em}
}

\author{
Tudor S\^ antejudean, Lucian Bu\c soniu
\thanks{The authors are with the Automation Department, Technical University of Cluj-Napoca, Romania. Email addresses: \texttt{tudorsantejudean@gmail.com}, \texttt{lucian@busoniu.net}. This work was been financially supported from H2020 SeaClear, a project that received funding from the European Union's Horizon 2020 research and innovation programme under grant agreement No 871295; by the Romanian National Authority for Scientific Research, CNCS-UEFISCDI, project number PN-III-P2-2.1-PTE-2019-0367; and by CNCS-UEFISCDI, SeaClear support project number PN-III-P3-3.6-H2020-2020-0060.}
}

\maketitle
\thispagestyle{empty}
\pagestyle{empty}

\begin{abstract}
We consider problems in which a mobile robot samples an unknown function defined over its operating space, so as to find a global optimum of this function. The path travelled by the robot matters, since it influences energy and time requirements. We consider a branch-and-bound algorithm called deterministic optimistic optimization, and extend it to the path-aware setting, obtaining \emph{path-aware optimistic optimization} (OOPA). In this new algorithm, the robot decides how to move next via an optimal control problem that maximizes the long-term impact of the robot trajectory on lowering the upper bound, weighted by bound and function values to focus the search on the optima. An online version of value iteration is used to solve an approximate version of this optimal control problem. OOPA is evaluated in extensive experiments in two dimensions, where it does better than path-unaware and local-optimization baselines.
\end{abstract}

% -------------------------------------------------------------------------
% ----- SECTION BREAK -------------------------------
% ------------------------------
\section{Introduction}\label{sec:intro}

We design and evaluate a method for a mobile robot to sample an unknown function defined over its operating area or volume, so as to find the global optimum of this function. The key difference from classical optimization is that the path taken by the robot is important, since it influences energy and time costs; and the function is unknown and must be learned from samples. We call this scenario ``path-aware global optimization''. It can be useful in many practical scenarios, where the optimum sought could be e.g.\ of some physical measurement such as pollutant concentration, temperature, humidity etc. \cite{Essa_2020, Lilienthal_2004}, the maximal density of surface or underwater ocean litter, the largest-bandwidth location for radio transmission \cite{Fink_2010}, the largest sand height on the seabed for dredging, maximal or minimal forest density in inaccessible areas \cite{Sankey_2017}, and so on. As a first approximation, the length of the path serves as a proxy for the costs, although of course more accurate models are possible that take into account the dynamics of the robot, the terrain etc.

Local optimization methods like gradient descent, which iteratively update a single point, can solve path-aware optimization after being modified to approximate derivatives from samples, as done e.g. in zeroth-order optimization \cite{Liu_2020}. We do evaluate such an approximate-gradient alternative; of course, the fundamental limitation is that these methods can only find a local optimum.

% Zeroth-order optimization is a gradient-free technique that applies the following $3$-step iterative process to converge to the optima: estimates the gradient from samples, computes the descent direction of it and updates the solution point .

\begin{figure}[!htb]
  \centering
  \includegraphics[width=0.8\columnwidth]{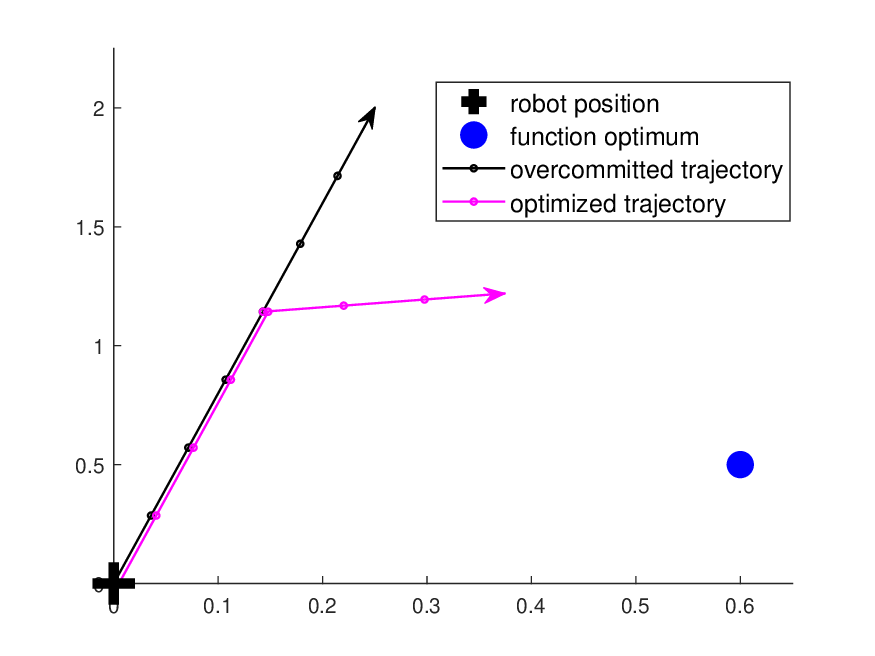}
  \vspace{-0.35em}
  \caption{Illustration of overcommitment.}\label{fig:over_comm_small}
  \vspace{-0.35em}
\end{figure}

% each cite take a textbook and maybe 2-3 representative refs. using e.g. https://arxiv.org/pdf/1812.03457.pdf
Global optimization techniques \cite{Horst_1996} on the other hand, like branch-and-bound, see \cite{Lawler_1996}, Chapter IV of \cite{Horst_1996}  or Bayesian optimization \cite{Frazier_2018, Shahriari_2016} usually make arbitrarily large ``jumps'' in the space of solutions to sample a new point. Here, large steps are unsuitable because they would overcommit: the robot samples the unknown function as it travels, so during a long path, new information becomes available, and the old travel direction might become suboptimal. Figure \ref{fig:over_comm_small} gives some intuition. Based on the information available to the robot when it is at the black cross, a classical algorithm decides to check the point at the black arrow. However, samples (dots) accumulated along the black trajectory give more information about the optimum (blue disk), so it becomes apparent that continuing along this trajectory would waste energy and time. Thus the robot changes heading to the magenta trajectory.

\vspace{-0.15em}

\vspace{-0.15em}

% Population-based techniques like genetic algorithms \cite{Mitchell_1998} or particle swarm optimization \cite{Poli_2007} would require large teams of robots, which are often unfeasible in practice.

In this paper, to solve the path-aware optimization problem, we extend an algorithm from the branch-and-bound class: deterministic optimistic optimization (DOO) \cite{Munos_2011, Munos_2014}. DOO organizes the search space as a tree containing at each depth a partition of the space, where the size of the subsets decreases with the depth. It makes a Lipschitz continuity assumption on the function and refines at each iteration a tree node with a maximal upper bound on the function value (i.e. a node that is likely to contain an optimum). We pick DOO because it guarantees a convergence rate towards the global optimum, which in essence says that only a ``subspace'' of a certain near-optimality dimension must be sampled to find the optimum \cite{Munos_2011}. % These are important features for path-aware optimization.

The main contribution of the paper is to make DOO path-aware, by formulating the decision of which direction to go at each time step as an optimal \emph{control} problem. Due to the danger of overcommitment, we do not immediately sample the largest-upper bound location, as DOO would, but we still exploit the underlying objective: to refine the upper bound around the optima. Specifically, the reward in the control problem is the volume by which each decision is expected to refine the upper bound, i.e. the difference between the bound before, and after measuring the new sample. This refinement is weighted by the value of the bound and of the function itself at the current point, to focus the refinement around the optima. We run an online version of a value iteration algorithm \cite{Dimitri_2019} to solve the control problem of maximizing the cumulative rewards (weighted refinements) along the trajectory. 

Solving the optimal control problem exactly at each decision step is impossible, both due to computational reasons and because finding the exact reward would require to know the future function samples. Therefore, we must resort to certain approximations, which we detail in the remainder of the paper. We validate the algorithm in extensive simulations, in which the key performance criterion is the path length taken to reach (close to) the optimum. We study the impact of the tuning parameters of the algorithm, its robustness to errors in the Lipschitz constant used to compute the upper bounds, and compare it to baselines adapted from classical DOO and gradient ascent.
%as well as illustrative real-time experiments. 

Related work can also be found in other fields than optimization. For example, in artificial intelligence, bandit algorithms are a class of sample-based optimization of stochastic functions \cite{Lattimore_2020}. They also overcommit by sampling at arbitrary distances, and must typically sample at least once everywhere to start building their estimates. In robotics, mapping requires building a map of the environment, and leads to the famous SLAM problem \cite{Durrant_2006} when the location of the robot is also unknown. Informative path planning chooses the path of a robot so as to find a map or other quantity of interest in as few steps as possible \cite{Binney_2010, Binney_2012, Lim_2015}. Other variants, like coverage \cite{Choset_2001}, aim to find a shortest path that examines the entire space using the robot sensors. Different from all these robotics paradigms, the aim here is not to build or sense the entire function, but rather just to find the optimum as quickly as possible.

% KEY TODO ITEM: Look for papers that solve similar problems. I am not finding anything...

Next, Section \ref{sec:Background} gives necessary background on DOO. Section \ref{sec:Problem statement} states the problem and formulates it as optimal control, while Section \ref{sec:Algorithm} provides the new algorithm to solve it. Section \ref{sec:Experiments} presents the numerical results, and Section \ref{sec:con} concludes the paper. 

% -------------------------------------------------------------------------
% ----- SECTION BREAK -------------------------------
% ------------------------------
\section{Background}\label{sec:Background}

% DOO -- reuse Automatica 14 removing SOO and the analysis stuff, reduce to 1 sentence: it has global opt and convergence rate guarantees

DOO is an algorithm belonging to the  branch-and-bound class that aims to estimate the optimum of a function $f:X\rightarrow  \mathbb{R}$ from a finite number of function evaluations. It sequentially splits the search space $X$ into smaller partitions and samples to expand further only those partitions associated with the highest upper bound values. After the numerical budget has been exhausted, the algorithm approximates the maximum as the state with the highest $f$ value evaluated so far. An assumption made by DOO is that there exists a (semi) metric over $X$, denoted by $l$, and $f$ is Lipschitz continuous w.r.t. this metric at least around its optima, in the sense:
\begin{equation}\label{eq:lipContinuity}
f(x^*) - f(x)\leq l(x^*,x), \forall x \in X
\end{equation}
where $x^* \in \mathrm{arg max}_{x\in X}f(x)$. Note that for convenience, we will require here the inequality property to hold for any pair $(x,y) \in X^2$:
\begin{equation}\label{eq:lipContinuityGen}
|f(x) - f(y)| \leq l(x,y), \forall x,y \in X
\end{equation}
and the Euclidean norm weighted by the Lipschitz constant will be chosen as the metric $l$ over $X$:
\begin{equation}\label{eq:lipConst}
l(x,y) = M||x-y||, \forall x,y \in X
\end{equation}
where $M$ represents the Lipschitz constant. The method can be extended to any metric $l$ obeying the assumptions present in \cite{Munos_2014}.

Here we will use an alternative approach to the partition splitting in DOO: the construction of a so-called ``saw-tooth'' upper bound \cite{Munos_2014}, defined as $B:X\rightarrow \mathbb{R}$ so that:
\begin{equation}\label{eq:uBFcn}
f(x) \leq B(x) = \min_{(x_s,f(x_s)) \in S} [f(x_s)+l(x,x_s)], \ \forall x \in X
\end{equation}
where $x_s$ is a sampled point and $(x_s,f(x_s)) \in S$, denoting with $S$ the set of samples (function evaluations) considered while building $B$. At each iteration, the next state to sample is given by the formula:
\begin{equation}\label{eq:maxUBEq} 
x_+ \in \mathrm{arg max}_{x\in X}B(x).
\end{equation}
The algorithm iteratively samples points selected with equation (\ref{eq:maxUBEq}). Note that $B$ is lowered (refined) with each new sample gathered by the robot, implicitly via (\ref{eq:uBFcn}).
%and can be explicitly evaluated over a grid of points.

% -------------------------------------------------------------------------
% ----- SECTION BREAK -------------------------------
% ------------------------------
\section{Problem statement}\label{sec:Problem statement}

% global optimization problem with physical movement -- bad name, we need a good one!

%smallest amount of real-world distance covered (~energy) -- remotivate with the examples from the intro (briefly - 2 sentences)

%Formulating the problem as optimal control

%dynamics of the robot -- simple-integrator but with possibility to extend to arbitrarily complex dynamics

%reward function -- careful discussion 
%picture with area refined 

Given an unknown function $f: X \to \mathbb{R}$, a global optimum must be found in the least number of steps. Consider the maxima locations:
\begin{equation}\label{eq:maxEq} 
x^* \in \mathrm{arg max}_{x\in X}f(x)
\end{equation}
where $x$ represents a physical location in the space $X	\subset \mathbb{R}^p$.
No previous knowledge of the function is available to the robot and thus the function must be learnt from the samples taken across a single trajectory. The path travelled is important due to energy and time considerations often encountered in practical scenarios. Another constraint is that, due to limited velocity, the robot cannot sample at arbitrarily distant positions across the state space, being limited to neighboring ones.

%We identify here an optimal control problem (OCP), in which the objective is the minimization of the travelled distance until $f^* \myeq \mathrm{max}_{x\in X}f(x)$ is found, without employing any costs per step, and having also motion dynamics constraints.
%We identify here an optimal control problem (OCP), in which the objective is to minimize the real-world distance travelled by an robot until the maximum is found, having also motion dynamics constraints (the next states are limited to the neighbors of the current position).

% Lucian: just skip this y business entirely, it complicates unnecessarily. Say at the end that it can be extended and cite the comm rob paper where it is explained how.

The motion dynamics are described by the $p$-dimensional positions $x \in X$
%, extra motion related states $y \in Y \subseteq\mathbb{R}^{n_y}$ (accelerations, heading directions, etc.) 
and system inputs $u \in U \subset \mathbb{R}^{p}$. Most of the times $p \in \{2,3\}$. For simplicity, we consider simple integrator dynamics with the possibility of extension to more complex dynamics. Thus, the discrete-time dynamics are given by $g: X \times U \to X$:
\begin{equation}\label{eq:motiondynamics} 
%[p_{k+1};y_{k+1}] = g([p_k;y_k], u_k)
g(x_k, u_k) = x_k + u_k = x_{k+1} % used as transition fcn for DP
\end{equation}
where $k$ indexes the step of the considered trajectory.

The solution we propose is inspired by DOO as it builds and refines with each sample gathered the saw-tooth upper bound of the function. 
%DOO assumes the sampled function to be globally Lipschitz continuous (or at least around its maxima position). 
Unlike DOO, it cannot sample arbitrarily far in the search space (recall the dynamics constraints) and thus the classical approach of always sampling the point with the highest $B$-value is inappropriate. By following it, the robot would not be using the samples gathered until the target is reached, possibly overcommiting to a trajectory that meanwhile became suboptimal. To address this issue, we make the algorithm aware of its path by defining an optimal control problem (OCP) with a different goal. 

We start with a high-level intuition of this OCP, followed by the formal definitions. We do not sample directly the highest $B$-value points, but rather lower the upper bound around the optima to implicitly find $x^*$. 
Thus, we aim to maximize the refinements of the upper bound around points of interest that either: $a)$ have high upper bounds and optimistically can ``hide" maxima points, or $b)$ have high function values that can lead to a maximum. The classical learning dilemma between the exploration and exploitation tradeoff arises here too: encouraging $a)$ will lead to excessive refinements in untraveled regions, less focused around high-value function points (too high exploration), while encouraging $b)$, the robot will overly refine areas where high function values were sampled and visit less untraveled regions that can possibly contain maxima (too high exploitation).

\begin{figure}[!htb]
  \centering
  \includegraphics[width=1\columnwidth]{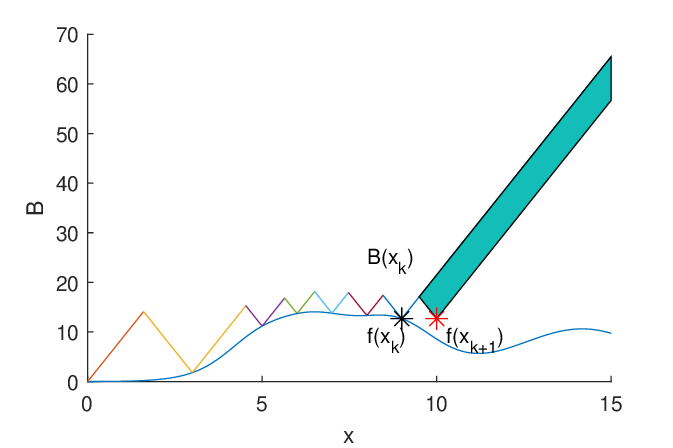}
  \vspace{-1em}
  \caption{The sampled function is drawn with a blue line and the saw-tooth envelope represents its upper bound. The function sample $f(x_k)$ (black star) is used to approximate $\widehat f(x_k)$, $\widehat f(x_{k+1})$ (red star) and determine $B(x_k)$. The area predicted to be refined is colored in green.}\label{fig:ref_plot_fin}
  \vspace{-1em}
\end{figure}

The OCP reward function defined next addresses this tradeoff. We consider the volume between the function upper bound and the horizontal plane (we use the term volume generically for any $p$; for $p=1$  it translates to area). With each new sample the function upper bound is lowered according to (\ref{eq:uBFcn}). The difference between the old and the new volumes is called volume refinement. Note that future samples are unknown and cannot be used to compute exactly this refinement and instead we must rely on approximations to predict it; refer to the example in Figure \ref{fig:ref_plot_fin} for more intuition. 

Formally, we define the reward as the volume predicted to be refined by taking action $u$ in state $x$, weighted by the average of the function value and its upper bound, both evaluated at $x$:

\begin{equation}\label{eq:rewFcn} 
\rho(x, u) = \frac{\widehat f(x)+B(x)}{2}r(x, u)
\end{equation}
where $\rho(x,u)$ is the reward function, $B$ is the upper bound function defined in (\ref{eq:uBFcn}) and $r(x, u)$ represents the volume predicted to be refined by taking action $u$ in state $x$. The refinement is computed in the following way. First, denote with $S=\{(x,f(x)) | x \in X\}$ the set of samples acquired so far. Compute next the upper bounds $B_1$ and $B_2$ using (\ref{eq:uBFcn}) and two slightly different sets of samples: $B_1$ with $S \cup \{(x,\widehat f(x))\}$ and $B_2$ with $S \cup \{(x,\widehat f(x)), (x_+,\widehat f(x_+))\}$, where $x_+=g(x,u)$. The volume refined is determined through trapezoidal numerical integration over the difference $B_1-B_2$ across the $p$ dimensions of $X$.

The terms $B(x)$ and $r(x, u)$ direct the refinements to locations with high upper bounds where optimistically a maximum might be situated (via $B(x)$) and where the robot has the potential to significantly lower $B$ (via $r(x, u)$).

Factor $\widehat f(x)$ in (\ref{eq:rewFcn}) tells the robot to visit states closer to those having high function values. Due to a limited number of samples acquired along the single-run trajectory, $\widehat f(x)$ is in most cases a prediction (especially at the beginning of the run). We compute this prediction by taking the function value of the closest point to $x$ that was already sampled. We do this for two reasons: taking a lower quantity than $\widehat f(x)$ would contradict the optimistic approach of our algorithm by overly encouraging exploitation, and taking a higher value, say closer to $B(x)$, would translate into an overly optimistic approach that encourages too much the exploration. If state $x$ was sampled before, its corresponding function value is directly taken. 

Figure \ref{fig:ref_plot_fin} gives an example of the reward calculation for a simple 1D case, where the evaluation point is denoted by $x_k$ and the next point (corresponding to $u_k$ in $\rho(x_k,u_k)$) is $x_{k+1}$. As $x_k$ was already sampled, its function value $f(x_k)$ is used to approximate $\widehat f(x_k)$, $\widehat f(x_{k+1})$ (as $x_k$ is the closest sampled point to $x_{k+1}$) and $B(x_k)$. The area predicted to be refined is colored in green and is of course an approximation, because one cannot guess in advance the next function samples, but only rely on approximations to predict them.

\vspace{-1em}

More generally, note that the ideal reward function would use true function values and updated $B$-values at the next steps. However, this is impossible in practice because doing so would require in advance knowledge of the function samples. This is why we must resort to approximations of $f$ and of the refinement.

Due to the online learning character, the OCP is a time-varying problem in which $\rho(x,u)$ changes with each new sample gathered as more information is available. % (so would the ideal $\rho$ because the $B$-values would change).
More specifically, the volume refinements $r$, function shape estimate $\hat{f}$ and the upper bound function $B$ change as the agent describes its path. This leads to a reward definition in (\ref{eq:rewFcn}) that is time-varying and we cannot derive the convergence guarantees present in classical value iteration. However, we expect the changes to be limited (since we only add one sample per step) and thus the solution of one problem should offer useful insight about the next. We exploit this feature in the algorithm from the next section. 

%+1: you need to explain that the ideal reward function would use true function values and updated B-values at the next steps. this is of course impossible in practice, because it would require knowing the function samples before we actually sample, so we MUST resort to approximations.

%Moreover, say that this is not just a simple OCP, but a time-varying one, because rho(x,u) changes at every step due to new information (so would the ideal rho because B-values would change). You should even give rho a time-step subindex, maybe. However, we expect changes are limited (one new sample at each step) so the solution of one problem offers useful info about the next, and we exploit this int he DP algo.

%explain clearly the differences from classical DOO
%In other words, the DP approach maximizes explicitly the upper bound based rewards, while the minimization of the travelled distance till maxima occurs as a consequence.

The OCP objective is to maximize the long-term value function $V:X\rightarrow \mathbb{R}$. As the horizon is unknown to the robot (there is no telling how many steps it will take to reach the maxima), we define $V$ in the infinite horizon setting:
\begin{equation}\label{eq:valFcn}
V^{h}(x) = \sum_{k = 0}^\infty \rho(x,u)
\end{equation}
where $h:X \rightarrow U$ represents the control law and $u = h(x)$. This objective aims to maximize the upper bound based rewards, and we will experimentally investigate the resulting travelled distance until the optimum.\footnote{We also tried to include an explicit travel cost term in the reward, but this did not work better.} We are searching for an optimal policy, denoted by $h^*$, such that:
\begin{equation}\label{eq:optRewObj}
V^{h^*}(x) \geq V^{h}(x), \forall x,h.
\end{equation}

Define also the optimal Q-function $Q^*(x,u) = \rho(x, u) + V^{h^*(x)}(g(x, u))$, which satisfies the Bellman equation:
\begin{equation}\label{eq:optQFcn}
Q^*(x,u)=\rho(x,u)+\max_{u'}Q^*(g(x,u),u')
\end{equation}
Once this equation has been solved to find $Q^*$, the optimal policy is given by $h^*(x) = \mathrm{arg}\max_u Q^*(x, u)$.

%The long-run maximization of the reward formula aims to lower the function envelope around high value function points to quickly find the maxima locations.

% best export EPS from MTB
%         saveas(figh, [savepath '.fig'], 'fig');
%         set(figh, 'PaperPositionMode', 'auto');
%         print(gcf, '-depsc2', '-loose', [savepath '.eps']);

% -------------------------------------------------------------------------
% ----- SECTION BREAK -------------------------------
% ------------------------------
\section{Algorithm}\label{sec:Algorithm}

% we solve the ocp def above using dynamic programming (DP) to approximate (\ref{eq:optQFcn}) ... and explain
Algorithm \ref{alg:OOPA} applies value iteration (VI) in an online scheme to solve the OCP defined above. %It aims to find an optimal path that refines the function upper bound in regions where it is large, and thus implicitly, like DOO, reaches close to the maxima. 
To build the upper bound and value function estimations, the robot needs to gather informative samples during its exploration procedure. Recall that samples can only be gathered in the current positions of the robot and are further used to approximate the function in unsampled points when computing the rewards in (\ref{eq:rewFcn}). 

%+2: you need more details on DP, not just the BEq. Give the update formula
%Q+ = rho(x,u) + ... Q ...
%and explain that you pick states and actions so that the next-state given by the dynamics always falls on the grid, so you only need to run the updates on the grid points x discrete actions x_i, u_j. Then say that many representation schemes can be used to get rid of this limitation, but for a first proof of concept we decided to not include them.

At each step $k$, the robot takes a sample $f(x_k)$ and adds it to the sample set $S$. Then, we update upon the current value function estimation by running $m$ value iteration (VI) updates of the following form:
\begin{equation}\label{eq:VI}
Q_+(x,u)=\rho(x,u)+\max_{u'}Q(g(x,u),u')
\end{equation}
based on the Bellman equation \eqref{eq:optQFcn}. For simplicity, the states are discretized into a grid $X_{grid}$, having $n_{grid}$ equidistant points along each of the $X$ dimensions. We pick states on $X_{grid}$ and actions so that the next state given the dynamics always falls on the grid. Actions leading the robot to the states up, down, left or right with one grid position with respect to the current position are considered. Thus, the VI updates \eqref{eq:VI} must be run only on states from $X_{grid}$ and on the discretized actions. Many representation schemes, including some for continuous actions, can be used to get rid of this limitation, and they will be studied in future work.

%+3 this is actually more complicated. Again, the true rewards would know HOW the B-values change (lower) with the new samples, meaning that the refinement gains at later steps would be smaller. Not knowing this means that rho necessarily overestimates at future steps, and that is the key reason for which you need to limit m. Moreover, the OCP also changes slightly at each step, see +1, which means you shouldn't cnoverge too close to the solution of the step-k OCP because at step-k+1 the OCp will change anyway.

% We are not trying to solve the OCP at every step, instead we only compute an estimation of $Q^*$, denoted by $Q$, using updates of the form: 
% \begin{equation}\label{eq:optQFcn}
% Q_+(x,u)=\rho(x,u)+\mathrm{max_{u'}}Q(g(x,u),u')
% \end{equation}

To fully solve the OCP, the true rewards should exploit how the $B$-values change (lower) with the new samples, meaning that the refinement gains at later steps would be smaller. Not knowing in advance the evolution of $B$ means that $\rho$ will overestimate these gains at future steps. This represents a key reason why $m$ needs to be limited. Moreover, the OCP changes slightly as it learns online and fully converging to the solution of step-$k$ would likely not be useful, since at step-$k+1$ the OCP will be updated again.
%The key reason is that the OCP changes slightly as new information becomes available at each step, which means that converging too close to a solution of the step-$k$ would overcommit as at step-$k+1$ the OCP will again be updated. Moreover, not knowing the true rewards in advance makes $\rho$ to unnecessarily overestimate at future steps, since the refinement gains decrease due to the lowering in the upper bound with each sample taken. This is the reason why $m$ needs to be limited. 

%This is due to the prohibitively expensive computations, and because applying these updates many times would extrapolate too much the overestimated rewards. The over estimations come from the unrefined regions with high $B$ values where the robot didn't yet sample. 
%Each reward requires the approximation of the volume refined, computed using the samples set $S$ and formula (\ref{eq:rewFcn}), as explained in Section \ref{sec:Problem statement}. 
The robot chooses next the action that maximizes the $Q$-values:
\begin{equation}\label{eq:maxRetAct}
u_k=\mathrm{arg}\max_u Q(x_k,u),
\end{equation}
applies it and measures the new state. The procedure continues until the number of steps in the trajectory, denoted by $n$, is exhausted.

Algorithm \ref{alg:OOPA} describes the steps presented above.

% define the OOPA algorithm. we need a better name for it
\newcommand{\algorithmicinput}{\textbf{Input:}}
\newcommand{\algorithmicoutput}{\textbf{Output:}}
\def\Input{\item[\algorithmicinput]}
\def\Output{\item[\algorithmicoutput]}

\vspace{-0.5em}
\begin{algorithm}[!htb]
    \caption{Path-Aware Optimistic Optimization (OOPA).} \label{alg:OOPA}
    \begin{algorithmic}[1]
    \Input $g$, $n_{grid}$ number of steps per $X$ and $B$ grid axes, discretized actions $U$, $m$ number of VI sweeps, $n$ trajectory steps, $M$ Lipschitz constant 
    \State generate $X$ and $B$ grid, $X_{grid}$, using $n_{grid}$
    \State initialize samples set $S \leftarrow \emptyset$
    \State initialize $Q_0(x, u) \leftarrow 0$, 	$\forall x \in X, u\in U$
    \State measure initial state $x_0$
    \For{each step $k = 0,\dotsc, n-1$}
        \State sample $f(x_k)$, add pair $(x_k, f(x_k))$ to $S$
        %\State $Q_0 = Q$
        \For{each VI sweep $m'=0, \dotsc, m-1$}
            \For{all states $x \in X_{grid}$, actions $u \in U$}
                \State $x_+ \leftarrow g(x, u)$
                \State find $\widehat f(x)$, $\widehat f(x_+)$ and $B(x)$ using $S$ and (\ref{eq:uBFcn})
                \State $S_1\leftarrow[S,(x,\widehat f(x))]$, $S_2\leftarrow[S_1,(x_+,\widehat f(x_+))]$
                \State compute $B_1(x), \forall x \in X_{grid}$ using $S_1$ and (\ref{eq:uBFcn}) 
                \State compute $B_2(x), \forall x \in X_{grid}$ using $S_2$ and (\ref{eq:uBFcn})
                \State compute $r(x,u)$ using trapezoidal integration 
                \State \qquad over $(B_1-B_2)$ across $X_{grid}$ 
                %\State compute $B(x)$ using (\ref{eq:uBFcn}) and $S$
                \State compute $\rho (x,u)$ using (\ref{eq:rewFcn})
                \State $Q_{m'+1}(x,u)=\rho (x,u) + \mathrm{max}_{u'} Q_{m'}(x_+, u')$
                \vspace{-1em}
            \label{ln:viupdate}
            \EndFor
        \EndFor
        \State $Q_0=Q_{m}$
        %\State $Q = Q_{m}$
        \State $u_k = \mathrm{arg max}_{u \in U} Q_{m}(x_k,u)$
        \State apply action $u_k$, measure next state $x_{k+1}$
%        \State $\theta_{k+1} = \theta_k + \alpha_k \varphi(x_k) [\max_u Q(x_k, u) - \widehat V(x_k; \theta_k) ]$
%            \label{ln:semigrad}
    \EndFor
\end{algorithmic}
\end{algorithm}
\vspace{-0.5em}

%tuning parameters
We call the algorithm Path-Aware Optimistic Optimization (OOPA). It has the following parameters that need to be tuned: $m$ representing the number of VI updates, the Lipschitz constant $M$ (used to compute $B$ in lines $12-13$ via (\ref{eq:uBFcn}) and (\ref{eq:lipConst})), and the discretization factor $n_{grid}$ that dictates the number of points taken across each dimension of $X$. The first parameter, $m$, impacts the propagation of the rewards across the state grid considered. Taking a too high $m$ will not only lead to high computation times that are not viable in practice, but also overly extrapolate the rewards that are mostly overestimations of their true quantities. Therefore, we recommend taking $m\leq5$. The Lipschitz constant is generally unknown and needs to be tuned empirically. Taking $M$ much lower than its true value will %directly impact the exploration and thus the underlying refinement objective
create an upper bound that no longer satisfies the inequality $f(x)<B(x)$, and thus the algorithm could break. A good approach is to take a rather high $M$ and lower it sequentially based on the feedback provided by the experiments. Finally, the grid should a priori be taken as large as feasible given the computational resources; we investigate the effect of the grid size in the next section. %The accuracy when searching for the maxima is, indeed, limited to $\frac{\delta}{2}$. 

% \begin{algorithm}[!t]
% \caption{OPC}
% \label{alg:opc}
% \begin{algorithmic}[1]
% \State \textbf{input:} state $\x_0$, model $\f$, $\R$, split factor $\M$, budget $\n$, Lipschitz constant $\bar{L}_{\val}$
% \State initialize $\Tree$ with root $0$ labeled by $\U^\infty$
% \While{computation $\n$ not exhausted}
% \State select box $\optim{i} = \argmax_{i \in \Leaves} \b(i)$
% \State select $\optim\k = \argmax_\k \disc^\k \d_{\optim i,k}$
% \State expand $\optim{i}$ along $\optim\k$: create its $M$ children on $\Tree$
% \EndWhile
% \State \textbf{output} sequence $\hat{\vect\u}$ of box $\opt{i} = \argmax_{i \in \Leaves} \val(i)$
% \end{algorithmic}
% \end{algorithm}

% -------------------------------------------------------------------------
% ----- SECTION BREAK -------------------------------
% ------------------------------
\section{Experiments and discussion}\label{sec:Experiments}

%% Simulation
At first, we define a standard setup in which we aim to study the new method and compare the OOPA algorithm against two baselines: Classical DOO (CDOO) that uses the saw-tooth approach to build the upper bound and commits to sample always the highest $B$-value point; and Gradient Ascent that creates an approximation plane using Local Linear Regression \cite{AtkesonMooreetal:97} on the neighboring samples and follows the plane's gradient to quickly converge to a local maximum.

A $21$x$21$ interpolation grid is taken across a state space of length $4$x$4$\,m. The function to optimize is composed of a sum of three radial-basis (RBF) functions with different coefficients: width $b_i \in \{[1.3; 1.3], [0.6; 0.6], [1; 1]]\}$, height $h_i \in \{[148.75, 255.0, 212.5]\}$ and centers $c_i \in \{[0.75; 1.5], [2.75; 3.5], [3.25; 0.75]\}$ (see Figure \ref{fig:e5_base_comp_comma_02} for a contour plot). So the global optimum is $f^*=255$ situated in $[2.75; 3.5]$. The corresponding Lipschitz constant is approximated starting from the Mean Value Theorem and then tuned experimentally to ensure that it produces close to true upper bounds. Thus, the Lipschitz constant was set to $M=364.54$. % The robot will sample this space to find as quickly as possible the maximum.

%The corresponding Lipschitz constant is approximated using the Mean Value Theorem on each of the 2D axes of $X$: for each separate RBF, the maximum absolute value of its derivative is computed. Then $M$ is set as the largest value between these maxima, obtaining $M=364.54$. The robot will sample this space to find as quickly as possible the maximum.

%% -------------------------------------------------------------------------
%% ----- SUBSECTION BREAK--------------------------------
\subsection{Influence of tuning parameters. Refinement prediction accuracy}\label{sub:Experiments:tuningParam}

The first parameter to be tuned is $m$, the number of VI updates at each step taken by the robot. On the setup defined above, we run the algorithm for $m\in\{1,2,3,4,5\}$ and study the maximum $f$-value sampled: $\overline{f}=\max_{x_s}(f(x_s))$, and the minimum distance until $x^*$: $\Delta_x=\min_{x_s}(||x^*-x_s||)$, where $x_s$ represents a sample point along the trajectory performed so far. Another metric of interest is the minimum difference between the optimum and the values of $f$ sampled so far: $\Delta_f=\min_{x_s}(f^*-f(x_s))$, with $x_s$ having the same meaning as above. The number of trajectory steps for each experiment is set to $n=125$, equivalent to $25$\,m travel distance. The robot starting position is set in the grid center, $x_0=[2;2]$.

\begin{figure}[!htb]
  \centering
  \includegraphics[width=1.0\columnwidth]{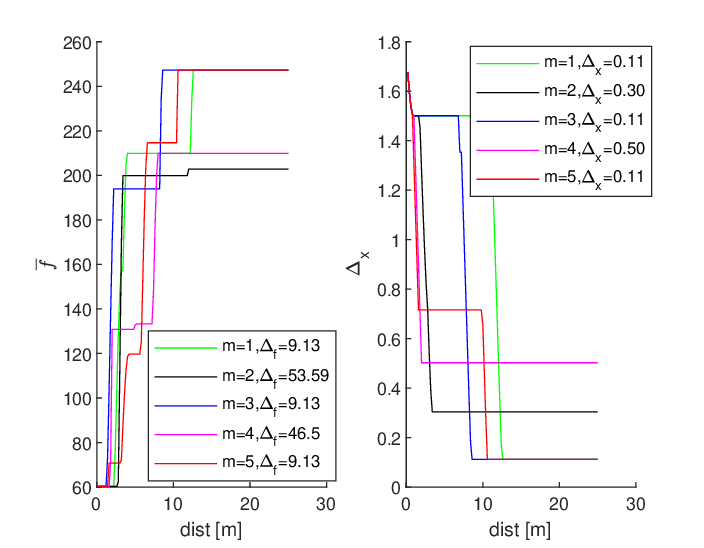}
  \caption{Illustration of the maximum value sampled so far (left) and the minimum distance to the maximum center $x^*$ (right), denoted with $\Delta_x$; for $m \in \{1,2,3,4,5\}$. The legend on the left shows the minimum distance between the samples of $f$ and its maximum value $f^*$, denoted by $\Delta_f$.}\label{fig:e2_m_comp_plot_no_title}
  \vspace{-1em}
\end{figure}

Figure \ref{fig:e2_m_comp_plot_no_title} shows that odd values of $m$ perform better in finding the maximum position with higher accuracy, %However, even values of $m$ do not find the maximum with accuracy of one grid step size (denoted by $\delta$) in the $n$ allocated, but still come as close as $N*\delta$ to it, where $N\in\{2,3\}$. According to Figure \ref{fig:e2_m_comp_plot_no_title} (left),
while even values of $m$ are suboptimal, converging to the second highest RBF.
%In practice, this limitation could be solved by applying some gradient ascent steps once a certain threshold has been reached. 
The best choice seems to be $m=3$, which gets close to the maximum faster and scores $20\%$ and $30\%$ less distance to $x^*$ compared to the cases of $m=1$ and $m=5$ ($8.4$\,m compared to $10.6$\,m and $12.4$\,m), respectively. The intuition is that rewards based on the volume refinements need to be propagated across the state space, however not too much, since they are mostly predictions (approximations) and in most cases their values are overestimated, especially at the beginning of the experiment. Choosing $m=3$ gives balanced results in terms of both exploration and exploitation.

\begin{figure}[!htb]
  \centering
  \includegraphics[width=1.045\columnwidth]{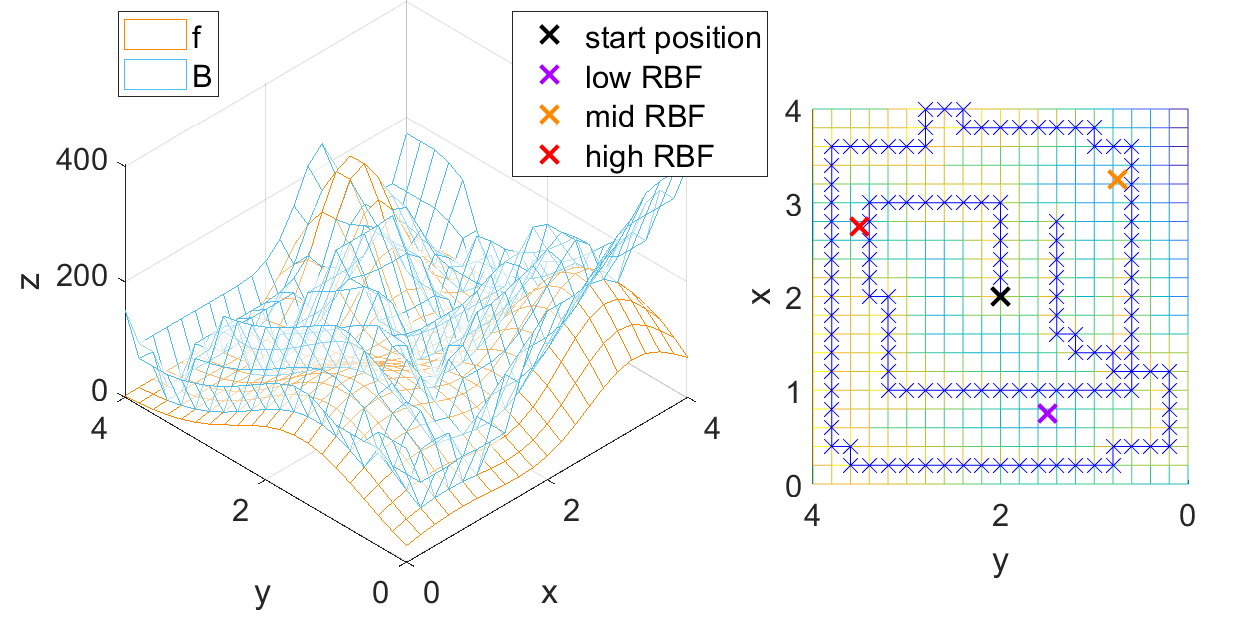}
  \vspace{-1.75em}
  \caption{Left: The sampled function (in orange) is bounded from above by the refined $B$ function (in blue), both evaluated on the same grid. Right: The sampling trajectory of the robot, drawn with blue 'x' and starting from the black 'x'. The refinements are geared -- without being overly committed -- towards higher function values (RBF peaks). This highlights a good exploration-exploitation tradeoff obtained for $m=3$.}\label{fig:e2_m3_end_ref_traj}
  \vspace{-2em}
\end{figure}

Figure \ref{fig:e2_m3_end_ref_traj} (left) shows the refined upper bound of the sampled function built using (\ref{eq:uBFcn}) and the $n=125$ samples acquired. The robot trajectory in Figure \ref{fig:e2_m3_end_ref_traj} (right) shows more steps being spent closer to the highest RBF (centered in $[2.75; 3.5]$) and less around the ones with lower values (centered in $[0.75; 1.5]$ and $[3.25; 0.75]$). This is expected, since the rewards take into account not only the upper bound, but also the function values.

Next, we will study the impact of the grid size on the behavior of the algorithm. For this, the state space will be split into grids of size $\{21^2,26^2,31^2,36^2,41^2\}$ and the behavior evaluated using the metrics from above. We keep the algorithm tuning and setup unchanged, with the exception of grid discretization and trajectory length set now to $75$m. This length is equal to the product between the grid step size and the samples budget $n$ for each experiment respectively.

\begin{figure}[!htb]
  \centering
  \includegraphics[width=1.0\columnwidth]{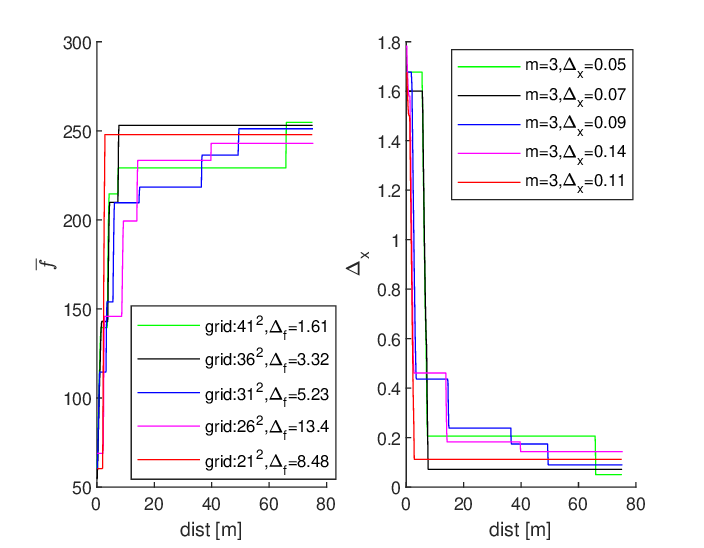}
  \caption{Smaller grid step sizes generally lead to better precision when searching for the optimum.  $\Delta_f$ is displayed in the legend on the left, while the legend on the right displays $\Delta_x$. }\label{fig:e3_grid_all_comp}
  \vspace{-1.5em}
\end{figure}

Except a ``lucky'' case ($21^2$), finer grids find the optimum with higher accuracy. Note that on all grids $x^*$ is found with a precision of $1$ grid step size,  as $m=3$ was kept constant during this experiment.

\vspace{-1em}
\begin{figure}[!htb]
  \centering
  \includegraphics[width=0.75\columnwidth]{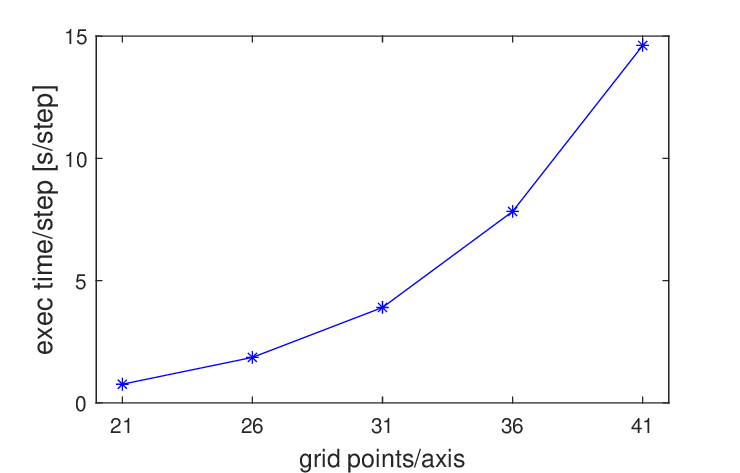}
  \caption{Time complexity has a quadratic growth with respect to the grid size.}\label{fig:e3_grid_all_tim_elp}
  \vspace{-1em}
\end{figure}

Time complexity has a quadratic growth with respect to the grid of points. Figure \ref{fig:e3_grid_all_tim_elp} shows the average execution time per step for each grid studied above.

Since in practice the Lipschitz constant is generally unknown, it is instructive to check the robustness of the algorithm in the event of overestimation or underestimation of $M$. Another goal is to provide a way of choosing $M$, as in practical cases computing the Lipschitz constant analytically is often impossible, and $M$ must instead be empirically tuned. We run the algorithm on the setup initially defined taking $M'=\lambda \cdot M$, with $\lambda\in\{0.2; 0.4; 0.6; 0.8; 1; 1.25; 1.5; 2; 2.5; 3\}$.

\begin{figure}[!htb]
  \centering
  \includegraphics[width=1.0\columnwidth]{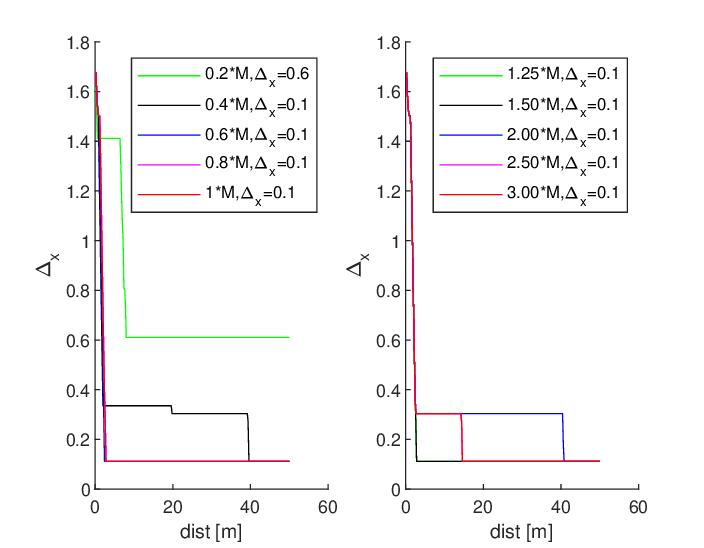}
  \caption{Robustness to underestimation (left) and to overestimation (right) of the Lipschitz constant. Taking $M$ less than half of its initial approximation leads to finding $x^*$ late or even breaks the algorithm. Taking $M$ more than twice of this approximation finds at worst later $x^*$.}\label{fig:e4_M_comp_simple_legend}
  \vspace{-1.5em}
\end{figure}

Figure \ref{fig:e4_M_comp_simple_legend} shows that taking $M$ lower that a half of its initial approximation can lead to finding the optimum late or even break the algorithm. On the other hand, overestimating the Lipschitz constant is a safer choice when its value is not (precisely) known. In this experiment, higher values of $M$ find $x^*$ at worst later and do not break the algorithm. A possible reason is that even though the volume refinements are weighted by the mean of the sampled function and its upper bound value, the latter has a greater impact due to its generally higher value. Decreasing $M$ too much will create an upper bound $B$ with much lower values compared to the true ones. So, as a rule of thumb, one should generally choose $M$ high and decrease its value based on experiments.

\begin{figure}[!htb]
  \centering
  \includegraphics[width=1.05\columnwidth]{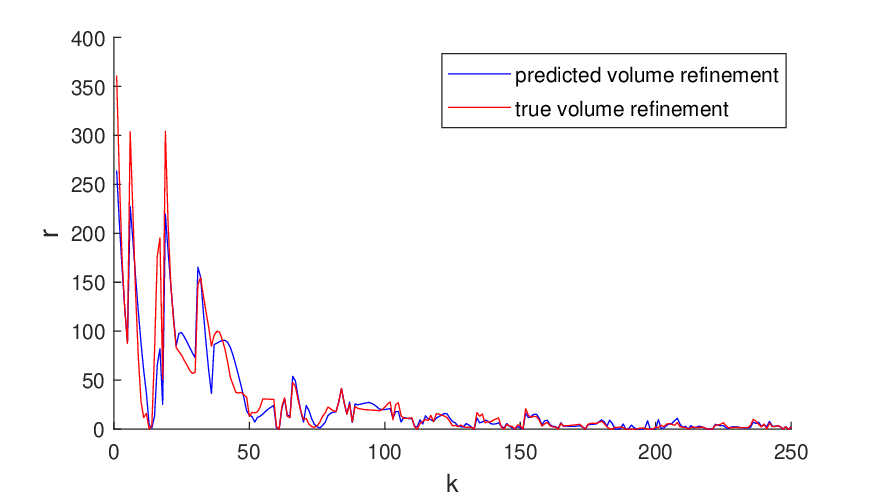}
  \caption{Illustration of accuracy of the upper bound refinements.}\label{fig:vol_ref_comp_250_compact}
  \vspace{-.5em}
\end{figure}

Finally, since reward values \eqref{eq:rewFcn} are based on predictions of upper bound refinements, we evaluate the accuracy of these predictions. Figure \ref{fig:vol_ref_comp_250_compact} compares at each step of the robot the predicted refinements with the actual ones (computed after the step has been executed and the new function value has been observed), for $m=3$ and $250$ trajectory steps. The two values follow largely the same trend, with somewhat poorer accuracy at the beginning of the simulation, improving as the upper bound decreases and the estimate $\widehat{f}(x_+)$ gets closer to the true $f(x_+)$. Thus, using the predictions is justified. 
% (recall how $r$ is computed in (\ref{eq:rewFcn})). Numerical results give an average mean squared error ($mse$) of 325 for all 250 trajectory steps compared to an $mse$ of just 6 for the second half (last 125 samples). This shows that the predicted values of the volume refinements converge (suboptimally) to their true quantities.

%% -------------------------------------------------------------------------
%% ----- SUBSECTION BREAK--------------------------------
\subsection{Comparison to baselines}\label{sub:Experiments:baselineComp}

The learning-based algorithm is next compared to two different baselines. The first one is represented by a classical DOO (CDOO) algorithm that, similarly to OOPA, applies the saw-tooth approach (\ref{eq:uBFcn}) to refine the function envelope with each sample taken. However, it fully commits to visit the maximum-$B$ point (\ref{eq:maxUBEq}) and thus changes its trajectory only once this point was reached. CDOO is not aware of the information gathered through the samples performed along the trajectory to the next target point.

The second baseline is Gradient Ascent. It applies Local Linear Regression on the closest $N$ neighbors to the current position to create a local approximation plane of the sampled function. This plane is differentiated and the gradient's direction that results is followed by the robot with maximum velocity. This method has the natural limitation of converging only to local maximum points, however at faster rates. 

To have a fair comparison, 15 equidistant starting positions placed along the lines drawn by the $3$ RBF centers (recall the initial setup) are considered. To give enough time to the algorithms to find the maximum, we set the number of steps per trajectory to $n=250$ ($50$\,m travel distance). The robot will move with a fixed step of $0.2$\,m (equal to the step size of the $21^2$ grid). Note that all methods sample the function with the same frequency, so that e.g. CDOO takes multiple samples along the (possibly overcommited) trajectory to the next chosen point. %, while CDOO and Gradient Ascent can change their step in the interval $[0;0.2]$\,m. 
For the gradient-based method $N=4$ ($N\geq3$ is required to build the approximation plane). 

\begin{figure}[!htb]
  \centering
  \includegraphics[width=1.05\columnwidth]{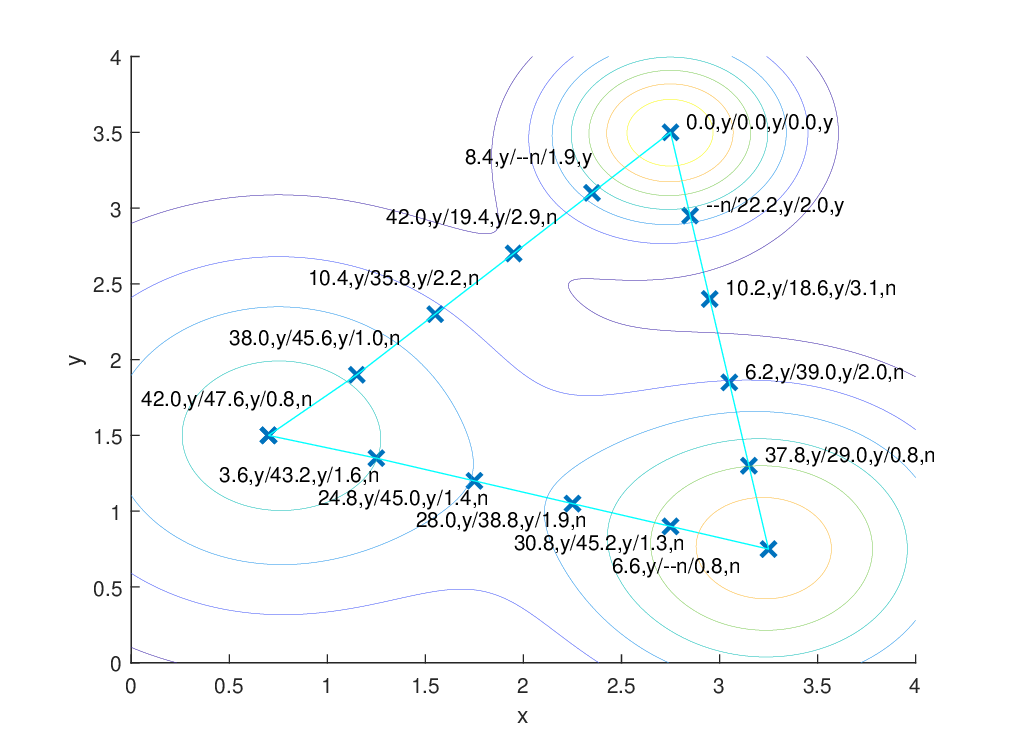}
  \vspace{-1.75em}
  \caption{Results of the OOPA, CDOO and Gradient Ascent methods.}\label{fig:e5_base_comp_comma_02}
  \vspace{-0.5em}
\end{figure}

We display in Figure \ref{fig:e5_base_comp_comma_02} the result of the experiments. Near each starting position, drawn with 'x', the distance until the optima for each method is given. The entries of each label correspond to OOPA, CDOO and Gradient Ascent, in this order, separated by slashes. For each algorithm, the number before the comma gives the distance traveled by the robot until convergence (when convergence does not occur, the distance is replaced by a `-'). After the comma, letters `y' (for yes) or `n' (no) show whether the maximum was found with accuracy of $1$ grid step size, $\delta=0.2$\,m. Note that for Gradient Ascent, we give the distance until convergence even when the algorithm reaches a local optimum (in which case 'n' is displayed). 

Excepting a few outliers, the DOO-based methods find the maximum position with accuracy $\delta=0.2$\,m ($1$ grid step size). When $x^*$ is found, OOPA scores $37.55\%$ less distance on average compared to CDOO. Gradient Ascent finds the maximum in only a fifth of the runs, mainly when it starts close to the center of the highest RBF. When this happens, the gradient-based method tends to find $x^*$ faster compared to OOPA or CDOO. In other initial states, Gradient Ascent converges to a local optimum. These behaviors are expected due to the local nature of the gradient and its straightforward choice of the heading direction. 
% Tuning experiment, comparing to baselines

% Resilience to overest/underest Lipschitz constant

% Resilience to multiple functions

% Illustration of resilience to noise (single-function)

% If gradient baseline, just do a one-para explanation on the spot

%% Real-time results, if they work

%%% -------------------------------------------------------------------------
%%% ----- SECTION BREAK--------------------------------
%%% ------------------------------
\section{Conclusions}\label{sec:con}

We considered the problem of finding a global optimum of a function defined over some physical space, by sampling it with a mobile robot. A method based on value iteration was defined to quickly reduce the upper bounds around optima, and thus implicitly find an optimum.

In future work we will consider more general robot dynamics than simple integrators, and demonstrate the method on real robots. Another objective is to provide near-optimality guarantees.

% \vspace*{-.5em}
\bibliographystyle{style/IEEEtranS}
%\IEEEtriggeratref{4}
\bibliography{main}

\end{document}